\documentclass[10pt,twocolumn,letterpaper]{article}

\usepackage{cvpr}
\usepackage{times}
\usepackage{epsfig}
\usepackage{graphicx}
\usepackage{amsmath}
\usepackage{amssymb}
\usepackage{multirow}
\usepackage{algorithm}
\usepackage[noend]{algpseudocode}
\usepackage{graphicx}
\usepackage{amsmath}
\usepackage{amssymb}
\usepackage{color}
\usepackage{soul}
\usepackage[table, dvipsnames]{xcolor}

\usepackage{enumitem}

\def\eg{\emph{e.g}}

\def\ie{\emph{i.e}}

\usepackage[pagebackref=true,breaklinks=true,letterpaper=true,colorlinks,bookmarks=false]{hyperref}

\cvprfinalcopy 

\def\cvprPaperID{3701} 

\begin{document}

\title{Exploring the Vulnerability of Single Shot Module in Object Detectors via Imperceptible Background Patches}


\author{Yuezun Li$^1$, Xiao Bian$^2$, Ming-ching Chang$^1$ and Siwei Lyu$^1$ \\
	$^1$ University at Albany, State University of New York, NY, USA \\
	$^2$ GE Global Research Center, Niskayuna, NY, USA}

\maketitle

\begin{abstract}

Recent works succeeded to generate adversarial perturbations on the entire image or the object of interests to corrupt CNN based object detectors.
In this paper, we focus on exploring the vulnerability of the Single Shot Module (SSM) commonly used in recent object detectors, by adding small perturbations to patches in the background outside the object. The SSM is referred to the Region Proposal Network used in a two-stage object detector or the single-stage object detector itself. The SSM is typically a fully convolutional neural network which generates output in a single forward pass. Due to the excessive convolutions used in SSM, the actual receptive field is larger than the object itself. As such, we propose a novel method to corrupt object detectors by generating imperceptible patches only in the background. Our method can find a few background patches for perturbation, which can effectively decrease true positives and dramatically increase false positives. Efficacy is demonstrated on 5 two-stage object detectors and 8 single-stage object detectors on the MS COCO 2014 dataset. Results indicate that perturbations with small distortions outside the bounding box of object region can still severely damage the detection performance. 
\end{abstract}

\vspace{-0.8cm}
\section{Introduction}

Convolutional Neural Networks (CNN) are shown to be vulnerable against {\em adversarial perturbations} \cite{goodfellow2014explaining}, which are intentionally designed and imperceptible noise added to the input that can drastically affect network performance. Many works \cite{szegedy2013intriguing,goodfellow2014explaining,kurakin2016adversarial,papernot2016limitations,moosavi2016deepfool,moosavi2017universal,zeng2017adversarial,luo2018towards,baluja2018learning,evtimov2018robust} have investigated this vulnerability and proposed various adversarial attack methods to impair image classifiers. 

\begin{figure}[t]
\vspace{0.5cm}
\centering
\includegraphics[width=0.95\linewidth]{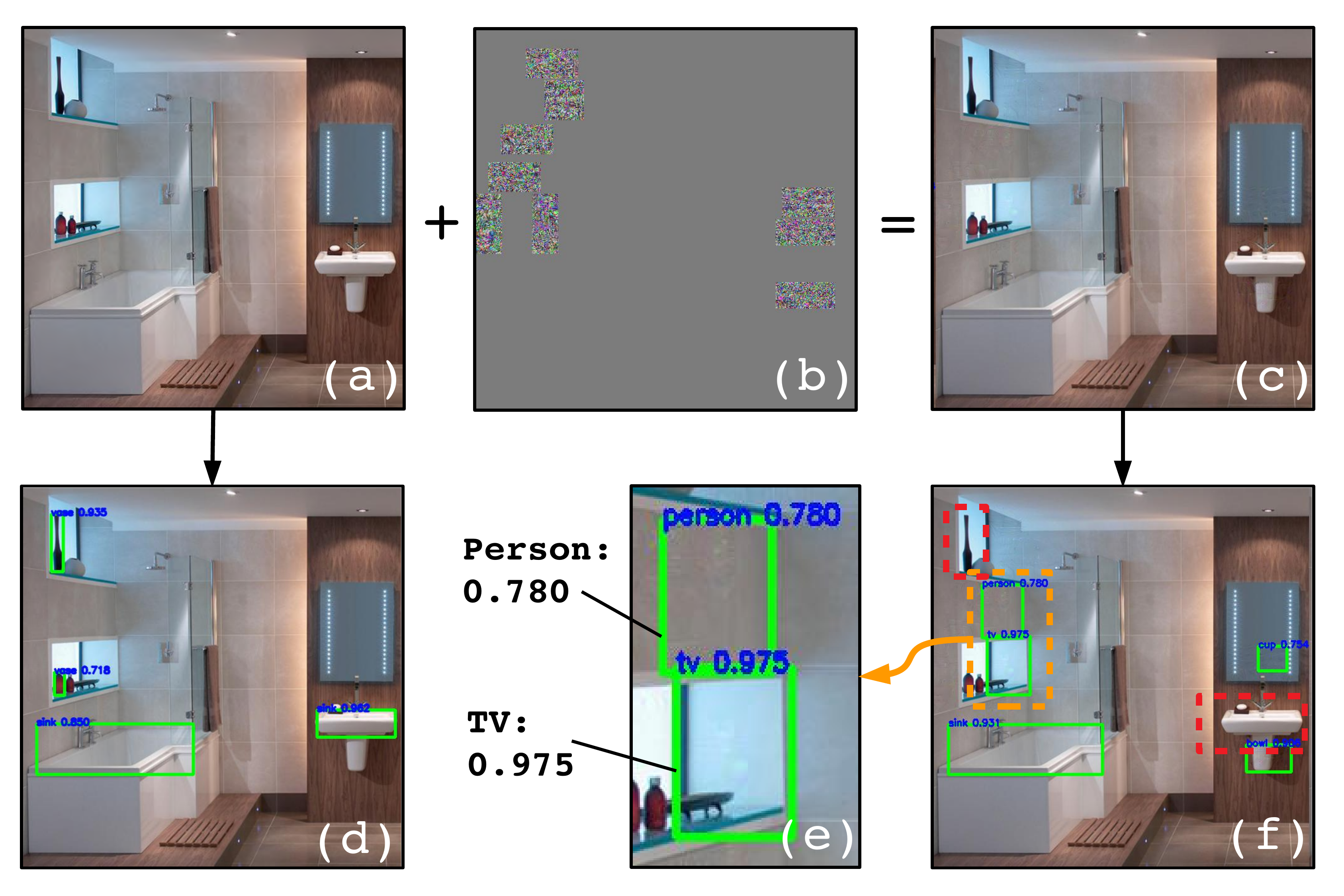}  
\caption{\small \em Visual illustration of the background patch attack on object detectors. (a) Original image. (b) Adversarial background patches,  amplified by a factor of 30 for visualization. (c) Perturbed image. (d, f) Detection results of (a, c) using Faster-RCNN \cite{faster-rcnn} respectively. (e) Zoom-in of the false positives ``person'' and ``TV'' in (d). Red boxes in (d) denote miss detections.}
\label{fig:tease}
\vspace{-0.2cm}
\end{figure}

\begin{figure*}[t]
\centering
\includegraphics[width=\linewidth]{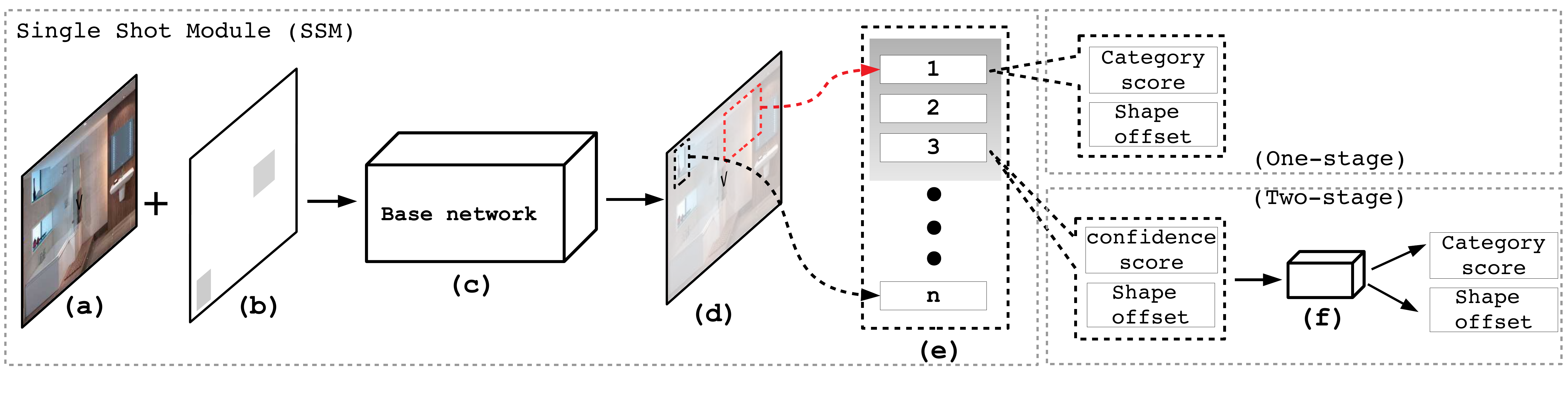}  
\vspace{-1cm}
\caption{\small \em Overview. (a) Original image. (b) Background patches generated by our method. (c) Base-network, which is the RPN for two-stage object detectors or the single-stage object detector itself. (d) Output of SSM, where the red box and black box denote a false positive and a true positive, respectively. 
Our attack can disrupt the top ranked results by decreasing true positives and increasing false positives. 
(e) denotes the top ranked results, which are the object proposals for two-stage object detectors or the detections for single-stage object detectors.
(f) Sub-network of two-stage object detectors for class labels prediction and shape refinements.
}
\label{fig:overview}
\vspace{-0.5cm}
\end{figure*}

Recently, adversarial perturbations are extended to networks for other computer vision tasks
such as the object detectors and semantic/instance segmentation networks \cite{lu2017adversarial,xie2017adversarial,chen2018robust,lu2017standard,eykholt2018physical,li2018rap}. 
However, all existing methods focus on creating adversarial perturbations on either the entire image or the object itself. An intuitive question to ask is: {\em can adversarial perturbations be added solely on the background to achieve similar vulnerability?} We will address this very problem in this paper. 
Specifically, we explore the vulnerability of the single shot feedforward network in the state-of-the-art two-stage \cite{faster-rcnn} and single-stage object detectors \cite{redmon2016you,liu2016ssd,liu2017receptive,li2017fssd} and show that the mechanism can be corrupted by adding imperceptible perturbations on a few small background patches, which can not only decrease true positives of detected objects, but also increase false positives in the background.
Figure \ref{fig:tease} shows a visual illustration of this approach. Looking forward, how to address the vulnerability of modern object detectors explored in this work will be a critical open issue, as applications including autonomous driving and AI medical image analysis demand highly reliable and trustworthy object detectors \cite{akhtar2018threat}.

Mainstream CNN based object detectors call into two categories: {\em two-stage} and {\em single-stage} object detectors. The {\bf Single Shot Module (SSM)} refers to the Region Proposal Network (RPN) in the two-stage object detectors or the single-stage object detector itself. Since the SSM makes use of excessive convolutions in multiple layers, the receptive field is often much larger than the object size.  This is precisely how contextual information outside the objects can be leveraged to improve detection.
However, this property also makes SSM vulnerable for attacks coming from the background.
If SSM was corrupted, no correct object proposals or detections will be generated, therefore leading to large errors.


In this paper, we propose a novel method to generate adversarial background patches to attack SSM.
Our method finds effective locations and shape for background patches to create adversarial perturbations inside, which  not only decreases the true positives, but also dramatically increases the false positives in the background.
To the best of our knowledge, most existing works focused on disrupting the true positives, and very few address the false positives --- which, in our view, is equally important regarding vulnerability in practical use, see Figure \ref{fig:tease}(e). Our method can effectively corrupt the SSM output ranking, such that false positives can be pushed ahead of true positives, see Figure \ref{fig:overview}(d-e).
%
%
Our adversarial background patch attack aims at achieving the following three aspects:
(1) decreasing the classification scores of correct detections, 
(2) corrupting the shape offset regression which shifts the localization (shape and location) of the correct detections, and 
(3) increasing the (non-background) object class scores that should not come up in the background. 
Our background patch attack generation can be cast as an optimization problem by minimizing the combination of the following three loss terms: 
(1) {\em True Positive Class ({\bf TPC}) loss}, which characterizes the correct class scores of the true positives; 
(2) {\em True Positive Shape ({\bf TPS}) loss}, which represents the correctness of shape offset regression of the true positives; 
(3) {\em False Positive Class ({\bf FPC}) loss}, which characterizes the non-background class scores of false positive arisen from the background. 
The combination of these three loss terms can be minimized using iterative gradient descent, such that the desired quantities regarding adversarial background patches (pixels, location and shape) can be calculated.


To demonstrate the efficacy of our method,
we conduct experiments on the MS COCO 2014 dataset \cite{lin2014microsoft} by attacking 5 mainstream two-stage object detectors and 8 single-stage object detectors. We conducted ablation studies investigating the vulnerability effects with respect to the following factors, and show how they can affect the degraded performance: (1) distance between generated background patches and the object of interest, (2) scale of the object to be detected, and (3) distances between objects.

The contributions can be summarized as four-fold:
\begin{itemize}[leftmargin=8pt,itemsep=0.01pt,topsep=0pt] 
    \item To the best of our knowledge, we are the first to explore the vulnerability of two-stage and single-stage object detectors by adding imperceptible adversarial perturbations on small patches in background.
    \item Our background patch attack can effectively decrease true positives and increase false positives in the background.
    \item We conduct comprehensive experiments on mainstream object detectors (5 two-stage and and 8 single-stage ones) to expose their vulnerability.
    \item Our method can generate `targeted' false positives of a given class in the background ($\S$\ref{subsec:targeted-fp}), which can cause serious vulnerability, \eg, to force a autonomous driving detector to trigger false pedestrian detections.
\end{itemize}

\section{Related Works}
\vspace{-0.1cm}

\noindent{\bf Object Detectors} detect objects of interest in an image by producing bounding boxes and class labels. The state-of-the-art object detectors are deep neural network based, where the network architecture consists of either a single-stage forward pass \cite{redmon2016you,liu2016ssd,liu2017receptive,li2017fssd} or a two-stage pipeline \cite{rcnn,fast-rcnn,faster-rcnn}. 
The two-stage object detector first generates object proposals, and then predict class labels and refine the shapes and locations of the proposals. 
Faster-RCNN \cite{faster-rcnn} is the very recent two-stage object detector, which improves the detection efficiency by using a Region Proposal Network (RPN) to generate object proposals. 
The RPN is essentially a fully convolutional neural network (FCN), which 
generates all object proposals in a single forward pass.
First a set of anchor boxes are identified, then object proposals are generated by estimating the location and shape of each anchor box with confidence scores. The top ranked object proposals will be selected for further classification and refinement. RPNs are effective and widely used in current two-stage object detectors. 
In comparison, single-stage object detectors \cite{redmon2016you,liu2016ssd,liu2017receptive,li2017fssd} can be viewed as an upgraded version of RPNs, where all detections are generated in a single forward pass (without generating proposals). 
Instead of predicting the confidence scores of the object proposals in RPN, single-stage object detectors directly predict the classification score for each detection. 

\smallskip
\noindent{\bf Adversarial Attacks for Image Classifiers.} Adversarial perturbations are intentionally designed noises that are imperceptible to human observers, yet can seriously reduce the deep neural network performance if added to the input image. Many methods \cite{szegedy2013intriguing,goodfellow2014explaining,kurakin2016adversarial,papernot2016limitations,moosavi2016deepfool,moosavi2017universal,zeng2017adversarial,luo2018towards,baluja2018learning} have been proposed to impair image classifiers by adding adversarial perturbations on the entire image. 
Recently, \cite{evtimov2018robust,brown2017adversarial} show that adversarial attacks can be generated in the physical world, using printable ``stickers'' that can be put on the objects in the scene to fool image classifiers. These stickers are clearly identifiable from human eyes but not the machines. The work of \cite{karmon2018lavan} generates an artificial adversarial patch in the background that is notable to human eyes but can cripple machine image classification. In contract, our method provides more sophisticated attacks on object detection, and the perturbation is imperceptible to human eyes. 


\smallskip
\noindent{\bf Adversarial Attacks for Object Detectors.} 
Recent research also explores the vulnerability of object detectors with extended network architectures \cite{lu2017adversarial,xie2017adversarial,eykholt2018physical,chen2018robust,li2018rap}. 
Object detection is widely used in practical applications such as autonomous driving, thus the impact of attack {\em vs.} vulnerability is greater. 
In \cite{lu2017adversarial}, adversarial perturbations are added to the stop-sign and face images that can cripple their detections. The dense adversary generation in \cite{xie2017adversarial} iteratively impairs both object detection and semantic segmentation. 
A physical adversarial perturbation method is proposed in \cite{chen2018robust} to attack Faster-RCNN based stop-sign detector. 
The work \cite{eykholt2018physical} extends \cite{chen2018robust} to other detectors such as YOLO, in that a physical adversarial sticker can cripple the stop-sign detection. 
The robust adversarial perturbations in \cite{li2018rap} corrupts deep proposal-based object detectors and instance segmentation methods by attacking the RPN, thus only two-stage networks are attacked.
%
All aforementioned methods focus on adding adversarial perturbations on the entire image or the object itself.  
Moreover, these methods have no intention to attack on the aspect of increasing false positives, their goal is solely on decreasing the true positives. 
In contrast, our adversarial attack is more universal in that: 
(1) most modern object detectors including both single-stage and two-stage ones are covered, 
(2) both the true and false positives of the detection are explicitly impaired, and 
(3) the added perturbation is imperceptible and only on the small patches in background.

\section{Methods}
\label{sec:method}

Our method can generate imperceptible background patches, which can effectively damage mainstream CNN object detectors, by simultaneously decreasing true positives and increasing false positives in background. 
The true positives will be corrupted with decreased classification score and largely shifted localization (shape and location). 
The false positives will be exacerbated with increased (non-background) class scores that should not come up in the background.
Our approach can be formulated as the minimization of three loss terms: {\em True Positive Class ({\bf TPC}) loss}, {\em True Positive Shape ({\bf TPS}) loss}  and {\em False Positive Class ({\bf FPC}) loss}, which we will formally define in $\S$\ref{subsec:tp-conf-loss}-\ref{subsec:fp-conf-loss}. 
The loss minimization can be computed using iterative gradient descent in $\S$\ref{subsec:bkg-patch-generate}, where the variables of unknown are the pixel perturbations and localization (shape and location) of the background patches. 

Our attack method works in a `white-box' manner, \ie, we assume that it has access to network parameter for back-propagation gradient computation.


\subsection{Notations and Problem Formulation}
\label{subsec:notation}

Let $\cal I$ denote the input image, $\{ \bar{b}_i = (\bar{x}_i,\bar{y}_i,\bar{w}_i,\bar{h}_i), i=1, ..., N\}$ denote the $N$ ground truth bounding boxes $\{\bar{b}_i\}$ for the objects of interest in image $\cal I$, where $(\bar{x}_i,\bar{y}_i)$ are the box center coordinates, $(\bar{w}_i,\bar{h}_i)$ are their widths and heights, respectively. Let $\{0,...,C\}$ denote the set of class labels, where $0$ is the background class, and $C+1$ is the number of classes. Specifically, for two-stage object detectors, $C=1$ as the focus is to attack RPN, only the two classes of background/object need to be considered. 
\footnote{
Note that our method does not require the original ground truth labeling (used for detector training).
In practice, we can use the first test detection results as the ground truth to compute the three loss terms.
}

Let ${\cal F}$ denote the Single Shot Module (SSM) with fixed model parameters. 
Let ${\cal F}({\cal I}) = \{(s^c_j, b_j), j=1, ..., M\}$ denote the SSM results of either RPN object proposals or single-stage detections on image $\cal I$~\footnote{We denote object proposals as detections hereafter for simplicity.}, where $s^c_j$ denote the score of class $c$ after softmax, $b_j$ denote the bounding box of the $j$-th detection, and $M$ is the number of SSM proposals/detections. 
Let $b_j = ({x}_j,{y}_j,{w}_j,{h}_j)$, where $({x}_j,{y}_j)$ are the box center coordinates, and $({w}_j,{h}_j)$ are their widths and heights, respectively. Let ${\cal Q} = \{(\tilde{x}_k, \tilde{y}_k, \tilde{w}_k, \tilde{h}_k), k=1,...,K\}$ denote the adversarial background patches, with locations specified by $(\tilde{x}_k, \tilde{y}_k)$ and shapes specified by $(\tilde{w}_k, \tilde{h}_k)$; $K$ is the number of background patches that will be added for attack. Let ${\cal I} \odot {\cal Q}$ denote the masked pixel regions on image $\cal I$ specified by the background patches $\cal Q$.

Our goal is to generate background patches with small pixel changes that can disrupt SSM. The adversarial attack is then the search of both the background patch geometry (location, size and shape) and pixel changes to be altered. We formulate this optimization as the minimization of three loss terms: TPC denoted as $L_{tpc}$, TPS as $L_{shape}$, and FPC as $L_{fpc}$. 
To maximize adversarial attack effects, while minimizing any distortion to the input image $\cal I$, we control the amount of pixel change inside the background patches $\cal Q$, by employing the {\em Peak Signal-to-Noise Ratio} (PSNR), a widely used metric for human perception of image quality. Smaller distortion leads to higher PSNR value.
Specifically, the adversarial background patch can be produced by minimizing the following loss function {\em w.r.t.} ${\cal I} \odot {\cal Q}$, considering the location and shape of the background patches $\cal Q$ and the included pixel value ${\cal I} \odot {\cal Q}$ as variables: 
\begin{equation}
\begin{array}{ll}
\displaystyle
\min_{{\cal I} \odot {\cal Q}} \; 
\left\{ 
L_{tpc}({\cal I} \odot {\cal Q}; {\cal F}) + L_{shape}({\cal I} \odot {\cal Q}; {\cal F}) + \right. \\ 
\left.
\hspace{0.9cm} L_{fpc}({\cal I} \odot {\cal Q}; {\cal F}) 
\right\}, 
\; \textrm{s.t.} \; \textrm{PSNR}({\cal I} \odot {\cal Q}) \ge \epsilon,
\end{array}
\label{equ:total}
\end{equation}
where $\epsilon$ is the lower bound of PSNR. 
Compared to a recent work \cite{li2018rap} which creates adversarial perturbations on the entire image to decrease true positives, 
our method creates adversarial perturbations only in selected background patches and can largely increase false positives as well.



\subsection{True Positive Class (TPC) Loss}
\label{subsec:tp-conf-loss}


Our approach attacks detectors by only introducing changes in the background. Since the sum of all class scores is $1$, the attack of decreasing the score of the correct class $c$ can be alternatively achieved by increasing the score of another running up class $\hat{c}$, such that $s^{\hat{c}} > s^c$ invalidates a good detection. 
To make this attack most effective, we try to increase the score of the running up class $\hat{c}$ with the largest score $s^{\hat{c}}$ among all classes except $c$, \ie, $\hat{c} \in \{0,...,C\}/\{c\}$.
This running up detection selection considers the following criteria:
(1) detection $b_j$ with IoU overlapping with its ground truth box greater than threshold $0.5$,
and 
(2) class score $s^{c}_j$ greater than threshold $0.1$. 
Let $z_j = 1$ for the detection $b_j$ satisfying the above two criteria, and $z_j=0$ otherwise.
The TPC loss $L_{tpc}$ sums up the cross entropy of scores from the selected running up detections $z_j$ among all $M$ detections:
\begin{equation}
\begin{array}{ll}
L_{tpc}({\cal I} \odot {\cal Q}; {\cal F}) = - \sum_{j = 1}^M z_j \log (s^{\hat{c}}_j).
\end{array}
\label{equ:label-loss}
\end{equation}
Minimizing Eq.\eqref{equ:label-loss} 
increases score $s^{\hat{c}}$ from an incorrect class that can effectively invalidates true positives.


\subsection{True Positive Shape (TPS) Loss}
\label{subsec:shape-loss}

Shape regression is an important step to refine the localization of detections (or proposals), where the locations and shapes of anchor boxes are adjusted to match the corresponding ground truth boxes, by expressing the localization in terms of {\em offsets}. In general, CNN object detectors are vulnerable under attack at the shape regression step, even when the classification is functioning perfectly, as good detections will be pushed away from their desired locations.
The TPS loss is designed to push away the predicted localization from the correct ones.
Let $\Delta x_j, \Delta y_j, \Delta w_j, \Delta h_j$ denote the predicted offset in terms of object center and bounding box size. Let $\Delta \bar{x}_j, \Delta \bar{y}_j, \Delta \bar{w}_j, \Delta \bar{h}_j$ denote the true offset between the corresponding anchor boxes and ground truth boxes. The TPS loss $L_{shape}$ sums up the squared offset differences of selected true positives under the criteria $z_j$ defined in $\S$~\ref{subsec:tp-conf-loss} among all $M$ detections:
\begin{equation}
\begin{array}{ll}
L_{shape}({\cal I} \odot {\cal Q}; {\cal F}) = 
\exp \left[ -\sum_{j =1}^M z_j \; \cdot
\right.
\\ 
\hspace{0.5cm}
 \left(
(\Delta x_j - \Delta \bar{x}_j)^2+(\Delta y_j - \Delta \bar{y}_j)^2 
\right.
\\ 
\hspace{0.5cm}
\left.\left.
+ (\Delta w_j - \Delta \bar{w}_j)^2 + (\Delta h_j - \Delta \bar{h}_j)^2
\right)
\right].
\end{array}
\label{equ:offset-disturb}
\end{equation}

Minimizing Eq.\eqref{equ:offset-disturb} 
encourages pushing the predicted offsets $\Delta x_j, \Delta y_j, \Delta w_j, \Delta h_j$ away from the true offsets $\Delta \bar{x}_j, \Delta \bar{y}_j, \Delta \bar{w}_j, \Delta \bar{h}_j$, to corrupt the predicted localization of $b_j$.
Note that in contrast to \cite{li2018rap} which disrupts the shape offset regression by pushing the localization toward a large constant value, here we directly optimize against known ground truth values, which should be more effective.

\subsection{False Positive Class (FPC) Loss}
\label{subsec:fp-conf-loss}

We introduce FPC as a novel loss term to strengthen the attack that can corrupt detectors by increasing false positives in the background. 
In the case without attack, the background should only contain detections (or proposals) with high scores belonging to the background class. 
To make detectors generate false positives, the attack should make the score of an object class $c' \in \{1,...,C\}$ greater than that of the background class $0$, (\ie, $s^{c'} > s^0$), to push the incorrect detections ahead to the top. The red box in Figure \ref{fig:overview} (d-e) shows one such example. To make such attack most effective, we propose to try pushing forward the class instance $c'$ with the largest score among $\{1,...,C\}$ in the FPC loss design. 
Specifically, since the goal is to create false positives in the background, only detections $b_j$ satisfying the following conditions need to be considered: 
(1) detection $b_j$ without overlapping with any ground truth box (\ie, fully in the background), and 
(2) detection $b_j$ with IoU overlapping with the generated background patches $\cal Q$  greater than threshold $0.1$. 
Let $r_j = 1$ for the detection $b_j$ satisfying the above criteria, and $r_j=0$ otherwise.
The FPC loss $L_{fpc}$ sums up the cross entropy of the selected $r_j$ scores among all $M$ detections:
\begin{equation}
\begin{array}{ll}
L_{fpc}({\cal I} \odot {\cal Q}; {\cal F}) = -\sum_{j = 1}^M r_j \log (s^{c'}_j).
\end{array} 
\label{equ:fp-loss}
\end{equation}

Minimizing Eq.\eqref{equ:fp-loss} encourages to increase the score $s^{c'}_j$ of some incorrect object class in the background, which thereby creates false positives. 


\subsection{Background Patches Generation}
\label{subsec:bkg-patch-generate}

Our method generates and refines the adversarial background patches in the overall loss optimization iterations. We describes how the background patches are initialized, and how they can be expanded and refined, according to the framework in Eq.\eqref{equ:total} incorporating the three loss terms.
Direct minimization of the loss function in Eq.\eqref{equ:total} with respect to ${\cal I} \odot {\cal Q}$ is difficult. Thus, we employ a standard iterative optimization scheme using gradient descent.

In general, an object detector finds multiple objects, and the closer the objects are, the greater their receptive fields overlap. 
This suggests that a single adversarial background patch can corrupt the detection of multiple objects, as long as the background patch are close enough to them.
To select where best to put in background patches for most effective adversarial attack, we consider the spatial distribution of the objects and the potential locations, shapes, and sizes of background patches.
We start with clustering the objects of interest into groups based on their spatial distances within the image. For each group, we empirically generate $n_b=3$ background patches as initialization. 

Algorithm \ref{alg:background patches} lists the pseudo code of our adversarial image generation procedure. We first compute the gradient of the overall loss term {\em w.r.t.} ${\cal I}_t$ as $\; {\cal G}_t = $
\begin{equation}
\nabla_{{\cal I}_t} \left[
L_{tpc}({\cal I}_t; {\cal F}) + L_{shape}({\cal I}_t; {\cal F}) + L_{fpc}({\cal I}_t; {\cal F})
\right]
\label{eq:grad}
\end{equation}
where $t$ denote the iteration number.~\footnote{
Note that we omit the $\cal Q$ term from Eq.\eqref{equ:total} in this gradient formula, as ${\cal I}_t$ implies the ${\cal I} \odot {\cal Q}$ masking has already been performed in the iterations.
} 
We next describe how the background patches ${\cal Q}$ are initialized and updated.

\smallskip
\noindent{\bf Initial background patches:} We consider candidates of background patches for each targeted object, with size initialized to $0.2$ of the object size, and aspect ratios $(1, 0.67, 0.75, 1.5, 1.33)$. Sliding windows are used to select, for each object group, the best location and shape of the background patches ${\cal Q}_0$, according the the following criteria:
(1) The distance between background patch and objects should not be less than a threshold, as $0.2$ of the largest object box side. (2) The patch with largest sum of ${\cal G}_t$ gradient intensities is preferred. (3) No selected patches should overlap. 
Such patch selection repeats until $n_b$ background patches are obtained for each group.

\noindent{\bf Expanding background patches:} In the subsequent iterations $t > 0$, 
${\cal Q}_t$ expends from ${\cal Q}_{t-1}$ with a small stride ($0.02$ of the shorter side of $\cal I$) in one of the 4 possible directions (left, right, top, down).
The extension direction is determined by the one where ${\cal G}_t$ gradient intensity in ${\cal Q}_t$ increases the most. In our method the background patch only expands (no shrinking) in the iterations until termination. 

The adversarial image perturbations at iteration $t$ is denoted as ${\cal P}_t$, which can be calculated as the intersection of the gradient image ${\cal G}_t$ of the overall loss and the current background patches ${\cal Q}_t$, \ie, ${\cal P}_t = {\cal G}_t \odot {\cal Q}$.
A L2 normalized perturbation $\hat{{\cal P}}_t$ is then calculated to update the adversarial image ${\cal I}_t$, using scale parameter $\lambda=30$.
The adversarial perturbed image ${\cal I}_t$ is then clipped into $[0, 255]$. 

The optimization iteration continues until any of the following condition is reached: (1) maximal iteration $T=250$ is reached, (2) no true positive selection are available for TPS and TPS, \ie, $\sum_{j = 1}^M z_j = 0$ or (3) the RSNR$({\cal I} \odot {\cal Q})$ is less than a maximal image distortion threshold $\epsilon$. Since the PSNR in lossy image compression is typically between 30 to 50 dB \cite{telagarapu2011image}, we empirically set $\epsilon = 35$ dB for two-stage object detectors and $\epsilon=30$ dB for single-stage ones. 

\setlength{\textfloatsep}{1pt}

\begin{algorithm}[t]
\caption{\small \em Background Patch Generation}
\label{alg:background patches}
\small{
	\begin{algorithmic}[1]
		\Require {SSM model ${\cal F}$; input image $\cal I$; maximal iteration $T$}
		\State ${\cal I}_0 = {\cal I}, t = 0$
		\While{$t < T$ and $\sum^M_{j=1} z_j \neq 0$}
		\State ${\cal G}_t = \nabla_{{\cal I}_t} \left[ L_{tpc}({\cal I}_t; {\cal F}) + L_{shape}({\cal I}_t; {\cal F}) + L_{fpc}({\cal I}_t; {\cal F}) \right]$
		\If {$t = 0$}
		\State ${\cal Q}_0 \leftarrow$ initial background patches
		\Else 
		\State ${\cal Q}_t \leftarrow$ expanded background patches
		\EndIf
		\State ${\cal P}_t = {\cal G}_t \odot {\cal Q}_t$ \Comment{perturbation in background patches}
		\State $\hat{{\cal P}}_t = \frac{\lambda}{||{\cal P}_t||_2} \cdot {\cal P}_t$ \Comment{perturbation normalization}
		\State ${\cal I}_{t+1} = \textrm{clip}({\cal I}_t - \hat{{\cal P}}_t)$ \hspace{0.5cm} \Comment{update $\cal I$ with pixel clipping}
		\If {PSNR$({\cal I}_{t+1} \odot {\cal Q}_t) < \varepsilon$}
		\State break
		\EndIf
		\State $t = t + 1$
		\EndWhile
		\Ensure Adversarial perturbed image ${\cal I}_t$
	\end{algorithmic}}
\end{algorithm}

\section{Experiments}
\label{sec:exp}

We perform experimental evaluations of the proposed adversarial attacks on mainstream object detectors. $\S$\ref{subsec:details} describes details on attacking 5 two-stage object detectors and 8 single-stage ones. $\S$\ref{subsec:targeted-fp} describes the `targeted' false positives attack as a novel attacking scheme, where the user can specify a desired class be produced by the attacked detector. $\S$\ref{subsec:transferring} evaluates the transferring ability of the proposed attack method among common network architectures. $\S$\ref{sec:ablation} performs three ablation studies on major factors that can affect performance.


\subsection{Experimental Setup}

MS COCO 2014 dataset \cite{lin2014microsoft} is used to evaluate the performance of the background patch attack. It contains 80 object class and a background class. We randomly select 2000 images from MS COCO 2014 validation set for experiments. The detection performance
is evaluated using ``mean average precision'' (mAP) metric \cite{everingham2010pascal} at Intersection-over-Union (IoU) threshold $0.5$ and $0.7$.

\begin{table*}[t]
	\small
	\centering
	\begin{tabular}{|l | c| c| c| c| c| c|| c| c|}
		\hline
		             & \bf No Noise    &  \bf Random &  \bf TPC & \bf TPS & \bf TPC+TPS  & \bf FPC & \bf TPC+TPS+FPC \\
		\hline
		{\bf FR-v16} \cite{faster-rcnn} & 62.4/48.7  & 62.5/48.9 & 50.7/38.8    & 51.2/38.0   & 48.1/37.4  & 50.7/38.3 & \bf 41.9/32.7\\
		\hline		
	  	 {\bf FR-mn} & 46.1/32.9  & 46.4/32.9 & 31.6/22.5    & 34.6/21.2   & 32.3/22.8  & 36.0/26.4 & \bf 26.6/19.3\\
		\hline
	   {\bf FR-rn50} & 64.7/52.7  & 64.7/52.2 & 47.7/40.1    & 47.2/35.1   & 47.3/39.5  & 52.2/43.9 & \bf 39.8/33.4\\
		\hline
		{\bf FR-rn101} & 66.0/56.0  & 65.8/55.7 & 39.9/32.4  & 42.0/29.8  & 40.4/33.0  & 53.9/48.5 & \bf 36.2/31.2\\
		\hline
	  {\bf FR-rn152} & 70.0/60.0  & 69.1/58.9 & 38.3/30.1    & 42.3/27.2  & 37.2/27.7  & 56.0/48.3 & \bf 36.8/31.7\\
		\hline
		\hline

      {\bf SSD-rn50} \cite{liu2016ssd}       & 46.6/37.2     & 47.2/37.1  & 39.7/30.0    & 37.4/25.4 & 38.8/28.8 & 33.9/28.8 & \bf 27.9/20.9  \\
		\hline
	   {\bf SSD-v16}                         & 48.3/37.0     & 47.8/37.1  & 36.9/26.4    & 31.0/18.0 & 36.3/24.6 &  26.2/21.9 &  \bf 24.5/17.4 \\
		\hline
	  {\bf RFB-rn50} \cite{liu2017receptive} & 48.9/40.3     & 48.7/41.2  & 36.2/27.4    & 43.7/38.4 & 37.0/27.1 &  31.6/27.7  &  \bf 26.1/20.5 \\
		\hline
	   {\bf RFB-v16}                         & 48.3/37.9     & 46.5/37.3  & 32.9/23.8    & 32.3/19.2 & 32.3/22.5 &  30.2/25.1 &  \bf 26.0/19.4 \\
		\hline
	  {\bf YOLO2-mn} \cite{redmon2016you}    & 46.6/30.4     & 45.4/29.9  & 35.7/23.1    & 26.9/15.4 & 35.3/22.1 &  23.6/17.5  &  \bf 22.3/15.3 \\
		\hline
	  {\bf YOLO3-mn} \cite{redmon2018yolov3} & 49.0/36.0     & 49.6/36.5  & 40.0/28.5    & 33.9/20.7 & 39.5/26.1 &  32.7/25.5  &  \bf 33.3/21.8 \\
		\hline
	  {\bf FSSD-rn50} \cite{li2017fssd}      & 51.2/41.5     & 51.4/42.2  & 39.8/29.7    & 38.3/27.2 & 38.4/27.6 &  31.7/31.3  &  \bf 28.8/20.8 \\
		\hline		
	  {\bf FSSD-v16}                         & 54.0/44.2     & 53.9/43.5  & 38.5/28.4    & 31.8/19.8 & 38.0/26.8 &  35.4/31.0  &  \bf 33.5/24.1 \\
		\hline
	\end{tabular}
	\caption{\small \em Performance of background patches attack on 5 two-stage object detectors with 5 different Region Proposal Networks (RPNs) and 8 single-stage object detectors at mAP $0.5$ and $0.7$. {\bf No Noise} denotes the original performance without adding noise. {\bf Random} denotes the performance of adding random noises on patches. {\bf TPC}, {\bf TPS}, {\bf TPC+TPS}, {\bf FPC} and {\bf TPC+TPS+FPC} denote the performance using corresponding loss terms respectively. Lower value denotes better attacking performance.}
	\label{table:bkg-attack}
	\vspace{-0.2cm}
\end{table*}

\begin{figure*}[t]
	\centering
	\includegraphics[width=1\linewidth]{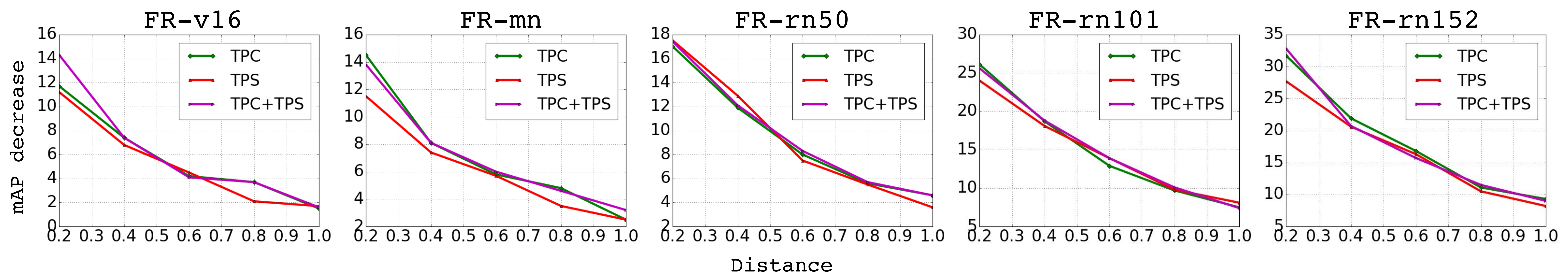}  
	\vspace{-0.5cm}
	\caption{\small \em Performance of background patch attacks at mAP 0.5 on 5 two-stage object detectors as the distance between background patches and objects (x-axis, normalized to $[0,1]$) increases. We do not consider FPC, as it is an independent factor here.
	}
	\label{fig:object-distance-plot}
	\vspace{-0.5cm}
\end{figure*}

\begin{figure*}[t]
	\centering
	\includegraphics[width=1\linewidth]{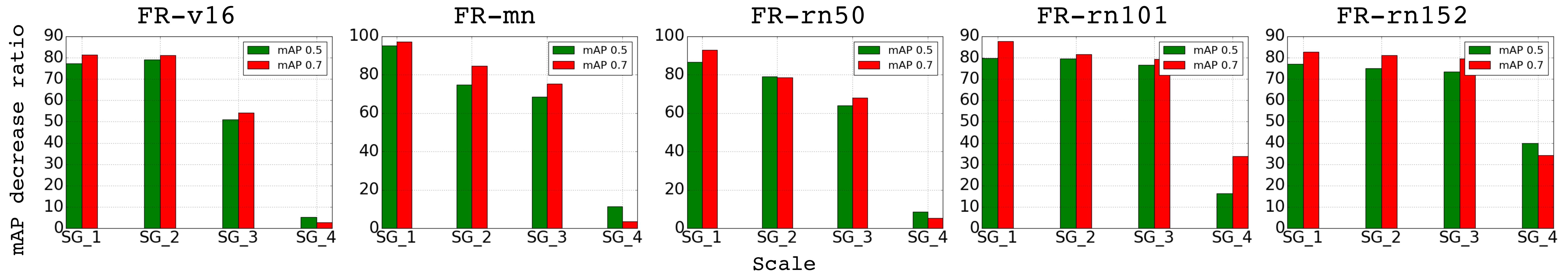}  
	\vspace{-0.5cm}
	\caption{\small \em Performance of background patch attacks at mAP 0.5 and mAP 0.7 at different object scales on 5 two-stage object detectors. SG\_1 - SG\_4 denote the 4 object groups sorted based on their size in ascending order.}
	\label{fig:object-scale-plot}
	\vspace{-0.3cm}
\end{figure*}

\label{subsec:details}
Experiments are conducted for 5 two-stage object detectors and 8 single-stage ones. For simplicity, we denote the base networks of these object detectors as: {\tt vgg16} {\bf (v16)} \cite{simonyan2014very} , {\tt mobilenet} {\bf (mn)}  \cite{howard2017mobilenets}, {\tt resnet50} {\bf (rn50)}, {\tt resnet101} {\bf (rn101)} and {\tt resnet152} {\bf (rn152)} \cite{he2016deep}. 

For two-stage object detectors, we tested Faster-RCNN ({\bf FR}) \cite{faster-rcnn} based on 5 different RPNs:~\footnote{As the RPNs use the same base network with detectors, we use the name of base network to denote different RPNs}  
{\bf FR-v16}, {\bf FR-mn}, {\bf FR-rn50}, {\bf FR-rn101} and {\bf FR-rn152}.
For single-stage object detectors, we tested {\bf SSD} \cite{liu2016ssd}, {\bf YOLO2} \cite{redmon2016you}, {\bf YOLO3} \cite{redmon2018yolov3}, {\bf RFB} \cite{liu2017receptive} and {\bf FSSD} \cite{li2017fssd} on different base networks: {\bf SSD-v16}, {\bf SSD-rn50}, {\bf RFB-v16}, {\bf RFB-rn50}, {\bf YOLO2-mn}, {\bf YOLO3-mn}, {\bf FSSD-v16} and {\bf FSSD-rn50}.

We evaluated five combinations of loss functions: (1) {\bf TPC}, (2) {\bf TPS}, (3) {\bf TPC+TPS}, (4) {\bf FPC} and (5) {\bf TPC+TPS+FPC}. We added a {\bf Random} baseline experiment for comparison, where random noise under normal distribution with the same distortion as in our method is added to the background patches. 

\begin{figure}[t]
	\centering
	\includegraphics[width=1\linewidth]{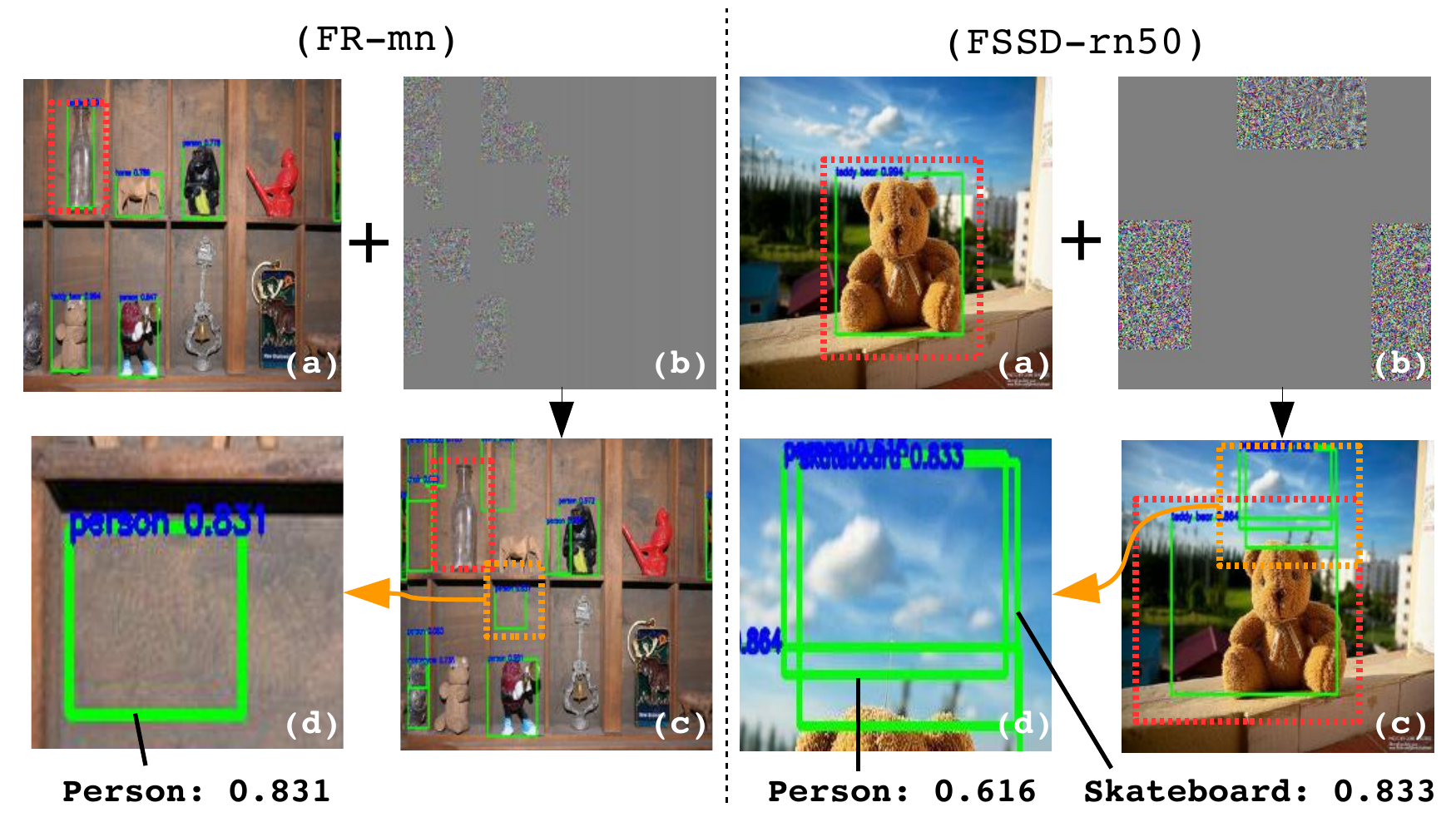}  
	\caption{\small \em Visual illustration of background patches attacks on {\bf FR-mn} and {\bf FSSD-rn50}. (a) Original detection results. (b) Background patches generated using TPC+TPS+FPC. (c) Detection results of adding (b) to the original image. Red boxes denote the true positives are not detected correctly. Specifically, the bottle in {\bf FR-mn} is not detected, and the bounding box of bear in {\bf FSSD-rn50} is shifted. Orange boxes in (c) denote the created false positives. (d) Zoom-in area in (c).}
	\label{fig:demo}
\end{figure}

Table \ref{table:bkg-attack} shows the detection performances under attack. Random noise barely affects the performance of object detectors. 
The performance decreases notably under the adversarial attacks. 
TPC+TPS considers both aspects to disrupt true positives.
We observe that the performance of TPC+TPS is in between TPC and TPS. Our explanation is that, since the amount of distortion is limited, minimizing the TPC+TPS loss could stop earlier than the cases only minimizing TPC or TPS individually. 
TPS is less effective in two-stage object detectors than the single-stage ones, as the shape offset attack can be mitigated by the later shape refinement in the sub-network of two-stage detectors. 
FPC largely reduces the performance of both kinds of  detectors, due to the increased false positives. 
The combined loss of TPC+TPS+FPC achieves the best performance, which decreases the performance by $\sim 42\%$ at mAP 0.5 and $\sim 40\%$ at mAP 0.7 for two-stage object detectors, $\sim 42\%$ at mAP 0.5 and $\sim 47\%$ at mAP 0.7 for single-stage detectors. 
Figure \ref{fig:demo} illustrates two examples of FR-mn  and FSSD-rn50 using overall loss TPC+TPS+FPC. The bottle (red box) in FR-mn is not detected, while a ``person'' (orange box) is yielded in background. The bear (red box) in FSSD-rn50 is classified correctly, yet the box shape is disturbed. Also, a ``person'' and a ``skateboard`` are yielded in the background.

\subsection{Targeted False Positives}
\label{subsec:targeted-fp}


The `targeted' false positives is an important adversarial attack scheme in practice, with the goal to
force a detector to mistakenly generate a false positive having specific given class. For example, to corrupt a traffic monitoring detectors to generate false positives of pedestrians in the background, that can greatly weaken the trustworthiness of the detector.
  
\begin{figure}[t]
	\centering
	\includegraphics[width=0.85\linewidth]{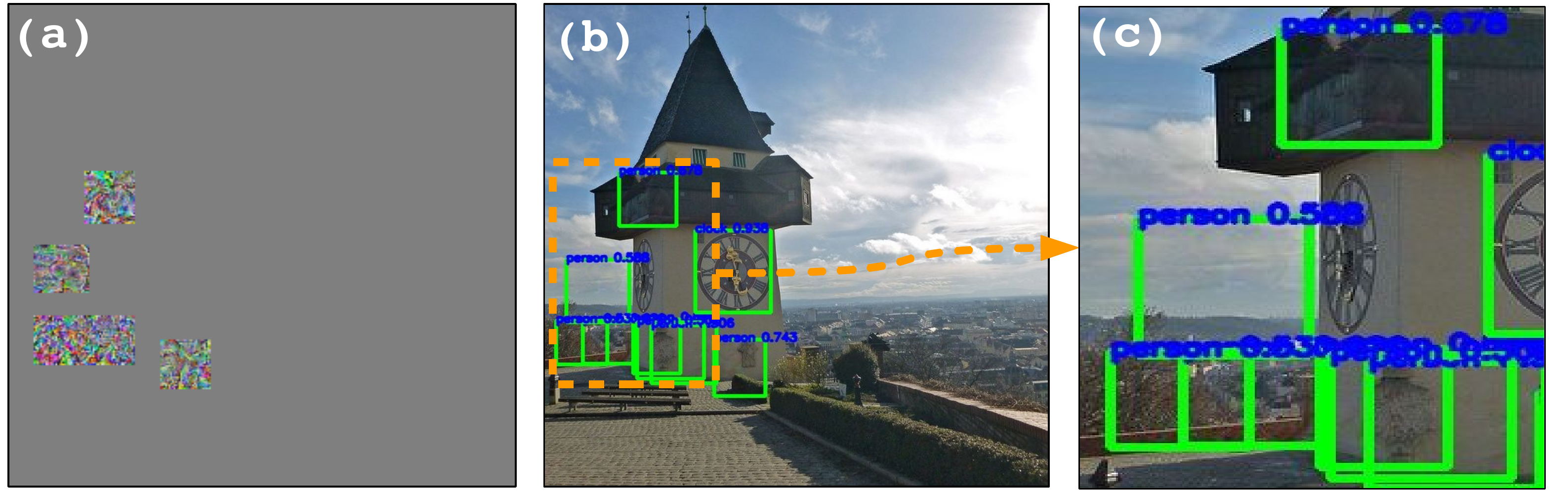}  
	\caption{\small \em Visual example of ``person'' targeted false positives on {\bf SSD-v16}. (a) Background patches generated using FPC loss function. (b) Detection result by adding (a). (c) Zoom-in illustration of ``person'' false positives.}
	\label{fig:targeted-fpc}
\end{figure}

Our method can generate targeted false positives using Eq.\eqref{equ:fp-loss} with a specified class, \eg, ``person''. Figure \ref{fig:targeted-fpc} illustrates an example of the ``person'' targeted false positives attack, where many ``person'' detections appear in the background. Table \ref{table:targeted-fp} shows the performance of the ``person'' targeted false positives attack on 8 single-stage object detectors, where the detection performance decreases by $\sim 8\%$ at mAP 0.5 and $\sim 9\%$ at mAP 0.7.
This indicates that targeted false positive attack is more challenging than an `untargeted' attack, where the performance largely drops by $\sim 38\%$ at mAP 0.5 and $\sim 32\%$ at mAP 0.7, as shown in Table \ref{table:bkg-attack}.

\begin{table}[t]
	\small
	\centering
        \begin{tabular}{|l | c| l| c|}
        \hline
		& \bf Targeted FPC & & \bf Target FPC\\
		\hline
	  {\bf SSD-rn50} & 44.7/35.0 & {\bf YOLO2-mn} & 38.7/26.0\\
		\hline
	   {\bf SSD-v16} & 46.4/35.1 & {\bf YOLO3-mn} & 44.8/32.3\\
		\hline
	  {\bf RFB-rn50} & 45.9/38.8 & {\bf FSSD-rn50} & 48.9/38.9 \\
		\hline
	   {\bf RFB-v16} & 43.9/34.2 & {\bf FSSD-v16} & 47.5/38.0 \\
		\hline
	\end{tabular}
	\caption{\small \em Performance of ``person'' false positives attack on the 8 single-stage object detectors.}
	\label{table:targeted-fp}
	\vspace{0.2cm}
\end{table}

\subsection{Transferring Study}
\label{subsec:transferring}


We study the transferring ability (how adversarial attacks generated on one detector can be used to attack another detector), to further explore the properties of vulnerability among common network architectures.
Table \ref{table:Transferring} reports the attack on 4 two-stage detectors and 4 single-stage ones. Our attack can transfer to similar network architectures. Background patches generated from FR-rn50 and FR-152 are more effective to FR-101 than FR-v16. Background patches generated from YOLO3-mn can also be effective to YOLO2-mn. The transferring ability between different base networks are weaker. Also, two-stage and single-stage object detectors can barely transfer attack from each other.

\begin{table*}[t]
	\small
	\centering
	\vspace{0.2cm}
	\begin{tabular}{|l | c| c| c| c|| c| c| c| c|}
		\hline
		 & {\bf FR-v16} & {\bf FR-rn50} & {\bf FR-rn101} & {\bf FR-rn152} & {\bf SSD-v16} & {\bf SSD-rn50} & {\bf YOLO2-mn} & {\bf YOLO3-mn}\\
		\hline
		{\bf No Noise}      & 62.4/48.7 & 64.7/52.7 & 66.0/56.0 & 70.0/60.0 & 48.3/37.0 & 46.6/37.2 & 46.6/30.4 & 49.0/36.0 \\
		\hline
		{\bf FR-v16}    & \bf 41.9/32.7 & 61.6/49.8 & 63.0/54.3 & 67.8/57.9 & 46.7/35.6 & 46.2/36.7 & 44.5/30.4 & 48.7/35.1\\
		\hline
		{\bf FR-rn50}   & 60.3/47.4 & \bf 39.8/33.4 & 62.0/53.6 & 67.6/57.1 & 47.8/36.0 & 46.4/36.5 & 45.5/30.2 & 48.5/35.2\\
		\hline
		{\bf FR-rn101}  & 62.2/48.0 & 60.7/49.6 & \bf 36.2/31.2 & 66.3/55.4 & 47.7/36.3 & 46.5/36.8 & 47.1/30.8 & 48.3/35.1\\
		\hline
		{\bf FR-rn152}  & 61.9/46.3 & 60.0/48.3 & 59.5/61.0 & \bf 36.8/31.7 & 47.6/36.1 & 46.8/36.8 & 44.8/30.3 & 48.8/35.3\\
		\hline
		\hline
		{\bf SSD-v16}   & 60.3/48.3 & 64.2/52.3 & 65.1/56.1 & 69.5/59.7 & \bf 24.5/17.4 & 45.4/36.4 & 46.5/30.5 & 48.3/36.2\\
		\hline
		{\bf SSD-rn50}  & 61.4/48.1 & 64.3/52.8 & 65.4/56.2 & 70.4/60.4 & 47.5/35.4 & \bf 27.9/20.9 & 46.5/30.4 & 49.1/36.1\\
		\hline
		{\bf YOLO2-mn}  & 61.6/48.8 & 64.4/52.4 & 65.6/56.7 & 69.7/59.6 & 47.8/35.8 & 46.6/36.9 & \bf 22.3/15.3 & 45.6/32.9\\
		\hline
		{\bf YOLO3-mn}  & 61.7/48.9 & 64.5/51.8 & 64.7/56.3 & 70.0/59.6 & 47.8/36.6 & 46.5/36.0 & 39.9/27.7 & \bf 33.3/21.3\\
		\hline
	\end{tabular}
	\caption{\small \em Performance of transferring attacks between 8 object detectors (4 two-stage and 4 single-stage). The row denotes where the background patches are generated from, and the column denotes each object detectors.}
	\label{table:Transferring}
	\vspace{-0.5cm}
\end{table*}

\subsection{Ablation Study}
\label{sec:ablation}

Ablation studies are performed to investigate the vulnerability of detectors under the proposed attack approach in the following three aspects that can affect performance.

\begin{figure}[t]

	\centering
	\includegraphics[width=1\linewidth]{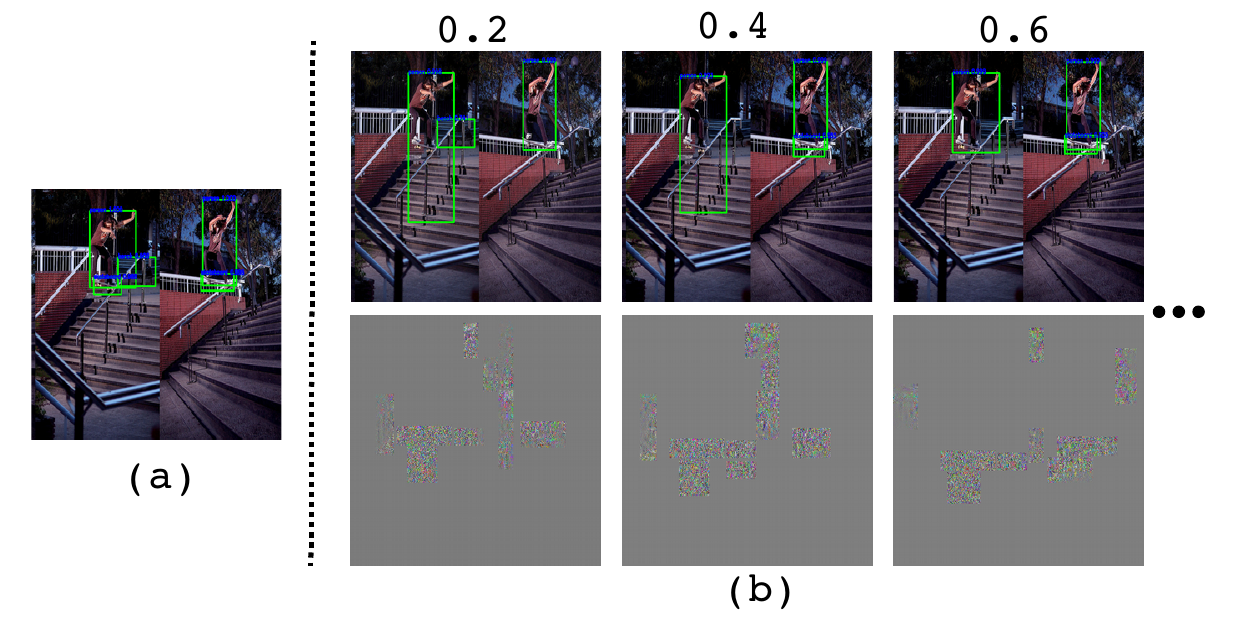}  
	\vspace{-0.6cm}
	\caption{\small \em (a) Original detection result of FR-rn152. (b) Detection result as the distance between background patches and objects increases (0.2-0.6). Person in the left is gradually detected as the distance increases.}
	\label{fig:object-distance-demo}
	\vspace{0.2cm}
\end{figure}

\smallskip
\noindent{\bf Distance between background patches and objects of interest:}
\label{subsec:distance}
We investigate the effectiveness of adding background patches at arbitrary location inside its theoretically receptive field. We normalize the distance between background patches and object to $[0, 1]$ by dividing the longest edge of object. We conduct experiments on 5 two-stage object detectors to show the detection performance decrease with distance between background patches and objects increases. Since FPC has no relation with the distance, we only consider three loss terms (TPC, TPS and TPC+TPS). Figure \ref{fig:object-distance-plot} illustrates the detection performance decrease as the distance between background patches and objects increases. We observe that larger distance, less performance decrease. Thus despite the receptive field is theoretically very large, the effectiveness of background patches attack gradually decreases as the distance increases, which is consistent with the explanation of receptive field in CNN \cite{zhou2014object}. We illustrate the background patches attack for FR-rn152 with distance increasing in Figure \ref{fig:object-distance-demo}. The left person in image is gradually detected out as the distance increases.

\smallskip
\noindent{\bf Object scales:} 
\label{subsec:object-scale}
We investigate how much object scales can affect the performance of the background patch attack. We cluster objects appeared in images into 4 groups, which is denoted by {\bf SG\_1 - SG\_4} sorted by their size in ascending order. Then we conduct experiments on 5 two-stage object detectors to show the attacking performance on each group. Table \ref{table:object-scale} shows the detection performance before and after background patches attack on different object scale. The first row of each object detectors denotes the original detection performance, the second row denotes the detection performance after background patches attack. For better visualization, we plot the detection performance decrease ratio in Figure \ref{table:object-scale}, y-axis is the detection performance decrease ratio. We observe that small objects are easier to be disturbed than large objects. This may be due to the fact that the convolution kernel size is fixed in each layer, so that the more context in surrounding background of small object can be seen compared to large object. 

\begin{table}[t]
	\small
	\centering
	\begin{tabular}{|l | c| c| c| c| c|}
		\hline
		             & \bf SG\_1    & \bf  SG\_2 & \bf SG\_3 & \bf SG\_4\\
		\hline
		\multirow{ 2}{*}{\bf FR-v16} & 19.5/14.4 & 29.2/24.3 & 35.5/31.9 &  64.1/54.4\\
		                             \cline{2-5}
		                             & \bf{4.4/2.7}   & \bf{6.1/4.6}   & \bf{17.4/14.6} &  \bf{60.7/52.9}  \\
		\hline
		\multirow{ 2}{*}{\bf FR-mn}  & 10.1/5.1 & 21.4/16.9 & 30.2/24.8 & 61.0/46.1 \\
		                             \cline{2-5}
		                             & \bf{0.7/0.4}  & \bf{5.4/4.6}   & \bf{9.5/6.1}   & \bf{54.1/44.5} \\
		\hline
	   \multirow{ 2}{*}{\bf FR-rn50} & 20.1/14.3 & 29.6/24.7 & 39.5/36.7 & 70.4/62.5 \\
		                             \cline{2-5}
		                             & \bf{2.7/1.0}   & \bf{6.2/5.1}   & \bf{14.2/11.7} & \bf{64.3/59.1}  \\
		\hline
	  \multirow{ 2}{*}{\bf FR-rn101} & 18.3/12.1 & 32.3/28.6 & 39.1/35.0 & 68.9/67.2 \\
		                             \cline{2-5}
		                             & \bf{3.7/1.5}   & \bf{6.6/5.3}   & \bf{9.1/7.2} & \bf{47.6/44.4} \\
		\hline
	  \multirow{ 2}{*}{\bf FR-rn152} & 20.0/15.7 & 30.8/26.6 & 41.7/38.3 & 72.0/69.6\\
		                             \cline{2-5}
		                             & \bf{4.6/2.7}   & \bf{7.7/5.0}   & \bf{11.1/7.8}  & \bf{43.2/35.7} \\
		\hline
	\end{tabular}
	\caption{\small \em Performance of 5 two-stage object detectors before and after (marked as bold) background patches attack on different object scale at mAP $0.5$ and $0.7$. SG\_1 - SG\_4 denote the 4 object groups sorted based on their size in ascending order.}
	\label{table:object-scale}
	\vspace{-0.05cm}
\end{table}

\smallskip
\noindent{\bf Distance between objects of interest:}
\label{subsec:distance-obj}
We observe a direct relation between the number of patches and the distance between objects. In particular, the smaller distances among objects, the fewer background patches are needed to successfully disturb the detections. We hypothesize it is also caused by the effect of receptive field. Specifically, when the objects are close to each other, their receptive fields also overlap, which implies that one patch can impact the detection of more than one object. We conduct experiments on 5 two-stage object detectors to validate our hypothesis. We normalize the mean object distance to [0, 1] by dividing the minimal dimension of image and we categorize all test images into 5 groups with different mean object distance and find out the mean number of background patches needed per object. As shown in Table \ref{table:patch-per-obj}, the number of patches per object reduces as the mean distance between objects in one image decreases. {\bf DG\_1 - DG\_5} denote the 5 groups which have mean object distance in a descending order. 

\begin{table}[t]
	\small
	\centering
	\begin{tabular}{|c| c| c| c| c|c|}
		\hline
		& \bf DG\_1    & \bf  DG\_2 & \bf DG\_3 & \bf DG\_4 & \bf DG\_5\\
		\hline
		{\bf Distance} & 0.55 & 0.30 & 0.23 & 0.18 & 0.09 \\
		\hline
		{\bf Patches}  & 3 & 1.79 & 1.49 & 1.23 & 1.22 \\
        \hline
	\end{tabular}
	\caption{\small \em Number of patches per object as the distance between objects decreases. DG\_1 - DG\_5 denote the 5 groups sorted based on distance between objects in descending order.}
	\label{table:patch-per-obj}
	\vspace{0.2cm}
\end{table}

\vspace{-0.2cm}
\section{Conclusion}
\vspace{-0.2cm}
In this paper, we explore the vulnerability of Single Shot Module (SSM) in mainstream object detectors, by adding imperceptible adversarial perturbations on small background patches outside the object. Our background patches attack can largely decrease the true positives and increase false positives in the background. Experiments on MS COCO 2014 dataset by attacking 5 two-stage object detectors and 8 single-stage ones demonstrate the efficacy. Future work includes the improvement of the optimization process and extension to attack black-box models.

{\small
\bibliographystyle{ieee}
\bibliography{ref}
}

\end{document}